\documentclass[runningheads]{llncs}
\usepackage[T1]{fontenc}
\usepackage{float}
\usepackage{amsmath} 
\usepackage{xcolor}
\usepackage{multirow}

\usepackage{algorithm}
\usepackage{algpseudocode}
\algrenewcommand\algorithmicrequire{\textbf{Input:}}
\algrenewcommand\algorithmicensure{\textbf{Output:}}

\usepackage{tikz}
\usetikzlibrary{shapes.geometric, arrows}

\tikzstyle{startstop} = [rectangle, rounded corners, minimum width=3cm, minimum height=1cm,text centered, draw=black, fill=red!30]
\tikzstyle{process} = [rectangle, minimum width=3cm, minimum height=1cm, text centered, draw=black, fill=orange!30]
\tikzstyle{arrow} = [thick,->,>=stealth]

\begin{document}

%
\title{Harnessing the Power of Semi-Structured Knowledge and LLMs with Triplet-Based Prefiltering for Question Answering}

%
\titlerunning{Triplet-Based Prefiltering for LLMs and Semi-Structured Knowledge}
%
\author{
Derian Boer\inst{1}\orcidID{0000-0002-2304-6613} \and
Fabian Koch \and
Stefan Kramer\inst{1}\orcidID{0000-0003-0136-2540}
}
\authorrunning{D. Boer, F. Koch, S. Kramer}
%
\institute{Institute of Computer Science, Johannes Gutenberg University Mainz, Staudingerweg 9,
55131 Mainz, Germany\\
\email{\{deboer, kramerst\}@uni-mainz.de}}
\maketitle              
\begin{abstract}
Large Language Models (LLMs) frequently lack domain-spe\-ci\-fic knowledge and even fine-tuned models tend to hallucinate. Hence, more reliable models that can include external knowledge are needed. We present a pipeline, 4StepFocus, and specifically a preprocessing step, that can substantially improve the answers of LLMs. This is achieved by providing guided access to external knowledge making use of the model's ability to capture relational context and conduct rudimentary reasoning by themselves. The method narrows down potentially correct answers by triplets-based searches in a semi-structured knowledge base in a direct, traceable fashion, before switching to latent representations for ranking those candidates based on unstructured data. This distinguishes it from related methods that are purely based on latent representations. 4StepFocus consists of the steps: 1) Triplet generation for extraction of relational data by an LLM, 2) substitution of variables in those triplets to narrow down answer candidates employing a knowledge graph, 3) sorting remaining candidates with a vector similarity search involving associated non-structured data, 4) reranking the best candidates by the LLM with background data provided.
Experiments on a medical, a product recommendation, and an academic paper search test set demonstrate that this approach is indeed a powerful augmentation. It not only adds relevant traceable background information from information retrieval, but also improves performance considerably in comparison to state-of-the-art methods. This paper presents a novel, largely unexplored direction and therefore provides a wide range of future work opportunities. Used source code is available at \url{https://github.com/kramerlab/4StepFocus}.

\end{abstract}

\section{Introduction}
\label{sec:intro}
Due to their increased capability to perform even complex tasks in recent years, LLMs are now being deployed extensively in day-to-day work, as well as in scientific and research contexts. To ensure accurate and reliable results, it is crucial to understand their abilities and limitations. LLM can sometimes provide incorrect answers, be unaware of gaps in their knowledge, or return biased information \cite{bender2021dangers}. Furthermore, they tend to ``hallucinate'' \cite{feldman2023trapping} in some cases, which can be challenging to detect. There is ongoing debate about their suitability for performing reasoning tasks \cite{subba24}.

\begin{figure}[t!]
	\includegraphics[width=\textwidth]{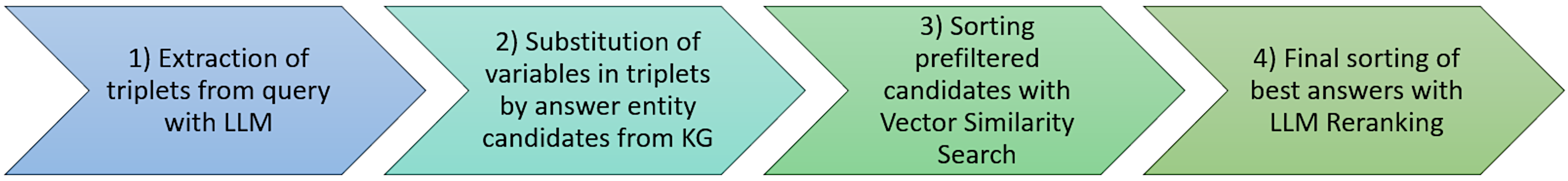}
	\caption{Pipeline of 4StepFocus that enhances VSS + LLM Reranker by triplet-based prefiltering steps.}
	\label{fig:framework}
\end{figure}

General LLMs may be used as \textit{foundation models}, that are customized for improved performance on specific tasks, languages, and contents. Methods such as transfer learning, dedicated prompt engineering, and enhancement with external data and knowledge are commonly employed. Wu \textit{et al.}~\cite{wu24} point out that external sources usually either include structured data, such as knowledge graphs (KGs), or unstructured documents. The authors recently assembled semi-structured knowledge bases (SKBs) and corresponding question-answer (QA) train and test data of three real-world domains 
in natural language, that require the combination of the included data. The six methods, which are evaluated on this benchmark datasets, are all based on word and graph embeddings
. They can be categorized into derivatives of vector similarity searches and graph neural network-based methods: both the dominating approaches in the LLM enhancement literature.
In contrast, we explore a novel and straightforward approach, 4StepFocus, to achieve more precise LLM responses whose selection increases traceability. Figure~\ref{fig:framework} visualizes the pipeline of our framework in short. 
It can be seen as a preprocessing step for approaches with vector similarity search, but involves having the LLM independently formalize the user input and search a KG for potential answers beforehand. We refer to a KG whose vertices are linked to documents with unstructured data, like user reviews, abstracts, or medical properties, as a semi-structured knowledge base (SKB) in this paper. 

Although further development and optimization are needed for better generalization and higher reliability, our experiments already demonstrate superiority over state-of-the-art methods. Importantly, in our method external information used in the decision process is more transparent and interpretable in a human-readable format. This ability to refer to quotable, semi-structured knowledge that is used in the reasoning process counters the current issue of LLMs being black boxes and makes their output more comprehensible for the user. 

The rest of this paper is structured as follows:
Section~\ref{sec:related} reviews related work on LLM enhancement for question answering. Section~\ref{sec:method} presents a detailed description of our approach. Section~\ref{sec:evaluation} describes the evaluation conducted on three QA testsets and compares our results to those of state-of-the-art methods. Section~\ref{sec:conclusion} summarizes our findings and outlines several options for future work.

\section{Related work}
\label{sec:related}

Our work builds on and extends the existing literature on integrating LLMs with KGs for question answering, which is related to various other applications such as fact-checking and hallucination mitigation. 
This section discusses several key contributions to these areas and situates our approach within this context. Due to the rapidity of new developments, we include not only peer-reviewed work but also preprints. 

Wu  \textit{et al.}~\cite{wu24} assembled three SKBs and created corresponding benchmark datasets for evaluating question answering. Their work includes a comparison of six methods:
\begin{itemize}
    \item \textit{Vector Similarity Search (VSS)} embeds both the query and the concatenated textual and relational information of each candidate entity. Subsequently, it computes the cosine similarity between the query and candidate embeddings. 
    \item \textit{Multi-Vector Similarity Search (Multi-VSS)} represents candidate entities with multiple vectors each to capture detailed features. In our case,
textual and relational information of each candidate entity are embedded separately.
    \item \textit{Dense Retriever} finetunes a query encoder and a document encoder separately using QA pairs from a training dataset.
    \item \textit{QA-GNN}~\cite{Yas21} constructs a subgraph where nodes represent entities found in the question or answer choices, incorporating their neighboring nodes. It integrates semantic embeddings from an LLM, jointly modeling both relational and semantic information.
    \item \textit{VSS + LLM Reranker}~\cite{Chia24,Zhu24} reranks the top-v results from VSS using LLMs. The LLM is given background information from the SKB about each of the top v results and prompted to return a score between 0 and 1, quantifying how much each of them fits as an answer to the question. Wu \textit{et al.} employed it with two different LLMs: GPT-4-turbo
(gpt-4-1106-preview) and Claude3 (claude-3-opus). 
\end{itemize}
We compare our results against this benchmark study to highlight the efficacy of our approach in Section~\ref{sec:evaluation}.

Much more related work has been done. We only mention a few, related ideas in the following:
Guan  \textit{et al.}~\cite{guan24} presented a method to mitigate LLM hallucinations via autonomous graph-based retrofitting. It is related to our approach, because it involves triplet generation as well, but still differs in some components, e.g., not involving VSS and not letting the LLM determine the triplets a priori.
Wang  \textit{et al.}~\cite{wang24} employ embeddings to answer questions spanning multiple documents.
FACE-KEG by Vedula and Parthasarathy~\cite{vedula21} uses a KG transformer network for explainable veracity prediction. 
Kundu and Nguyen~\cite{Kundu24} generate embeddings from a KG containing true claims and another with false claims, which are considered in the LLM's answer. 
The Knowledge Solver by Feng \textit{et al.}~\cite{feng2023knowledge} uses an LLM itself for pathfinding between identified and multiple-choice questions. 
Shakeel \textit{et al.}~\cite{shakeel21} describe a comprehensive pipeline using graph embeddings and various NLP techniques for fact-checking. 
Mountantonakis and Tzitzikas~\cite{mount23} provide a semi-automatic web application that assists users in comparing facts generated by ChatGPT with those in KGs by using multiple RDF KGs for enriching ChatGPT responses. 
Pal \textit{et al.}~\cite{pal2023medhalt} introduce the Med-HALT dataset and benchmark, focusing on testing LLMs in the medical domain through multiple-choice questions, confidence measures, and link generation. 
Liu  \textit{et al.}~\cite{liu2023context} train LLMs to generate SPARQL queries with annotated question-SPARQL pairs, facilitating more accurate KG-based question answering.
Li  \textit{et al.}~\cite{li24} present a workflow that includes the identification of main entities from queries, like our method, and augments the entities with further directly linked information from a KG. 
Unlike in our approach, unstructured data are not considered. 

Recent surveys~\cite{agrawal2023knowledge,pan24} 
provide comprehensive overviews of the integration of KGs with LLMs, discussing current challenges and future directions. Another study on the opportunities and challenges at the intersection of LLMs and KGs
~\cite{pan2023large} further explores the potential and obstacles in this interdisciplinary field.
Our work advances these lines of research by proposing a novel triplet-based prefiltering method for VSS and LLM Reranking that leverages SKBs, offering deeper contextual ``understanding'' and improved retrieval accuracy in question-answering.

\begin{algorithm} [t!]
\caption{4StepFocus}
\label{alg:4StepFocus}
\begin{algorithmic}[1]
\Require
    \Statex a semi-structured knowledge base $SKB = (V,E)$, a query $q$ in natural language, candidates for query answers $C\subseteq V$, maximum number of candidates to rerank $k_{max}$
\Ensure A list of most likely answers to $q$ in descending order
\State $T , x_{target} \gets \Call{ask\_llm\_for\_triplets}{q,V.types,E.types}$ \Comment{extracts triplets $T$}
\Comment{from $q$ and marks the variable that denotes the final answer to $q$ as $x_{target}$.}
\State $C_{filtered} \gets \Call{substitute}{T,x_{target},SKB}$
\State $C_{filtered}, C_{additional} \gets \Call{vss}{q, C_{filtered}}$
\State \Return $\Call{llm\_reranker}{q,C_{filtered}, C_{additional},k_{max}}$

\end{algorithmic}
\end{algorithm}

\begin{algorithm} [t!]
\caption{SUBSTITUTE}
\label{alg:Substitute}
\begin{algorithmic}[1]
\Require
    \Statex a set of raw triplets $T$, a target variable $x_{target}$, a semi-structured knowledge base $SKB$
\Ensure A set $C_{filtered}$ of candidate nodes for $x_{target}$, satisfying all triplets in $T$
    \State $T' \gets \Call{prepare\_triplets\_for\_substitution}{T}$
    \ForAll{$\tau_i=(h_i, e_i, t_i) \in T'$}
        \If{$h_i$ is a constant and $t_i$ is a variable}
        \State $substitute[t_i] \gets substitute[t_i] \cap neigbors(h_i, e_i, -)$
        \EndIf
        \If{$h_i$ is a variable and $t_i$ is a constant}
            \State $substitute[h_i] \gets substitute[h_i] \cap neigbors(t_i, e_i, +)$
        \EndIf
        \If{$h_i$ is a variable and $t_i$ is a different variable} 
            \State $N_i \gets \emptyset$.
            \ForAll{$h_i^{(j)} \in substitute[h_i]$}
                \State $N_i \gets N_i \cup neigbors(h_i^{(j)}, e_i, -)$
            \EndFor
            \State $substitute[h_i] \gets substitute[h_i] \cap N_i$
            \State $N_i \gets \emptyset$.
            \ForAll{$t_i^{(j)} \in substitute[t_i]$}
                \State $N_i \gets N_i \cup neigbors(t_i^{(j)}, e_i, +)$
            \EndFor
            \State $substitute[t_i] \gets substitute[t_i] \cap N_i$
        \EndIf
    \EndFor
    \State Repeat lines 2-21 until no changes occur anymore.
    \State \Return $substitute[x_{target}]$

\end{algorithmic}
\end{algorithm}

\begin{algorithm}
\caption{PREPARE\_TRIPLETS\_FOR\_SUBSTITUTION}
\label{alg:Prepare}
\begin{algorithmic}[1]
\Require
    \Statex a set of raw triplets $T$
\Ensure a set of valid, prepared triplets $T'$
    \ForAll{variables $x$ present in $T$}
        \State $substitute[x] \gets \bigcup_{type(c) = type(x)} c \forall c \in C$ \Comment{This step is not directly executed in the code implementation to save resources.}
    \EndFor
    \State $T' \gets \emptyset$ \Comment{Initialize a set for all valid triplets.}
    \ForAll{$\tau_i=(h_i, e_i, t_i) \in T$:}
        \If{$h_i$ is a constant and $t_i$ is a constant} skip $\tau_i$.
        \ElsIf{$h_i$ is a constant and $t_i$ is a variable}
            \If{$h_i$ (almost) matches any alias of any $v \in C$} 
                \State $T' \gets T'\cup \tau$ with $h_i$ replaced by the matching actual node $v$.
            \EndIf
        \ElsIf{$h_i$ is a variable and $t_i$ is a constant}
            \If{$t_i$ (almost) matches any alias of any $v \in C$} 
                \State $T' \gets T'\cup \tau$ with $t_i$ replaced by the matching actual node $v$.
            \EndIf
        \ElsIf{$h_i$ is a variable and $t_i$ is a different variable}
            \State $T' \gets T'\cup \tau$  
        \EndIf
    \EndFor
\end{algorithmic}
\end{algorithm}

\section{4StepFocus}
\label{sec:method}

We introduce a pipeline to enhance an LLM by an SKB in a novel, traceable fashion. A semi-structured knowledge base consists of a knowledge graph $G=(V,E)$ and associated text documents $D=\bigcup_{v \in V} D_v$ \cite{wu24}, where $V$ is a set of graph nodes and $E \subseteq V \times V$ a set of their connecting edges, which may be directed or undirected, and weighted or unweighted. 

While Figure~\ref{fig:framework} visualizes the pipeline of our framework in short,  Algorithm~\ref{alg:4StepFocus} in combination with Algorithms~\ref{alg:Substitute} and \ref{alg:Prepare} describe it in more detail.

We define the following helper functions for the algorithms:
    \begin{itemize}
        \item $type(v)$ returns the node type of any node $v \in V$. $V.types=\bigcup_{v \in V} type(v)$ denotes the set of all existing node types in SKB.
        \item $type(x)$ returns the node type $\upsilon \in V.types$ of any variable $x$.
        \item $type(e)$ returns the edge type of any edge $e \in E$. $E.types=\bigcup_{e \in E} type(e)$ denotes the set of all existing edge types in SKB.
        \item $neigbors(v, e', +)$ denotes a set of all nodes $n \in V$ that are connected to $v$ via an edge of type $e' \in E.types$ pointing from $n$ to $v$.
        \item $neigbors(v, e', -)$ denotes a set of all nodes $n \in V$ that are connected to $v$ via an edge of type $e' \in E.types$ pointing from $v$ to $n$.
    \end{itemize}

Algorithm~\ref{alg:4StepFocus} implements the four steps listed in Figure~\ref{fig:framework}: In line 1, function ASK\_LLM\_FOR\_TIPLETS($q,V.types,E.types$) prompts the LLM to return a sequence of triplets $T$ which formalizes $q$. Each triplet consists of a head entity $h_i$, a tail entity $t_i$, and a connecting edge type $e_i \in E.types$: $T = ((h_1, e_1, t_1), (h_2, e_2,  t_2), ..., (h_{|T|}, e_{|T|}, t_{|T|}))$.
Each $h_i$ and $t_i$ needs to be either a constant, i.e., a specific semantic entity, or a variable, i.e., representing an unknown semantic entity of the type $\upsilon \in V.types$. Additionally, the LLM returns a reference to the target variable $x_{target}$. Line 2 calls 
Algorithm~\ref{alg:Substitute}, in which the variables in $T$ are substituted by filtered answer candidates $C_{filtered}$ that satisfy the valid triplets. (Line 1 in Algorithm~\ref{alg:Substitute} in turn calls Algorithm~\ref{alg:Prepare}, which checks the validity of triplets and searches constants in $T$ in the SKB.) Finally, lines 3 und 4 of the main algorithm 
call the vector similarity search (VSS) and then LLM\_RERANKER, which computes the final rainking of $C_{filtered}$. Both VSS and LLM\_RERANKER \cite{wu24} have been briefly described in the Related Work section. $k_{max}$ denotes the number of those elements, presorted by VSS, that LLM\_RE\-RANKER finally ranks. If $|C_{filtered}| < k_{max}$, $k_{max}-|C_{filtered}| = |C_{additional}|$ elements are added to the solutions in our approach, although $(V,E)$ do not support their relevance to $q$. Hence, candidates in $C_{filtered}$ remain ranked higher.

\section{Evaluation}
\label{sec:evaluation}
We evaluate our method on the three STaRK benchmark datasets~\cite{wu24} and associated SKBs \textit{AMAZON}, \textit{MAG}, and \textit{PRIME}. While the medical SKB \textit{PRIME} with 10 entity types and 18 relation types has the highest proportion of relational data, STaRK-AMAZON has only 4 node types but a high proportion of unstructured data (i.e., user reviews). Each dataset comes with a subset of manually designed QA pairs and a set of QA pairs that have been automatically created with LLMs included in the mining procedure. 
The SKBs contain several million nodes and relations. In \textit{PRIME}, all nodes are marked as potential answer candidates, in the other cases it is only a high proportion. In most cases, there is more than one correct answer. 
We refer to Wu  \textit{et al.}~\cite{wu24} for a detailed analysis of the data's properties.
We use GPT4-o (\textit{gpt-4o-2024-05-13}) as the LLM for 4StepFocus 
and the embedding model \textit{text-embedding-ada-002} for VSS. We set $k_{max}$ of LLM\_Reranker to 20 in order to limit the amount of LLM requests. In the helper function $neigbors(v, e', -)$, we ignore the edge type while finding neighbors, because restricting it turned out to be error-prone in our experiments. 
We refer to Section~\ref{sec:related} and Wu \textit{et al.}~\cite{wu24} for details about all other benchmark methods and their configurations.

The following evaluation metrics were used and averaged over all queries. \textit{Hit@k:} 1 if a correct item is among the top-k results from the model, 0 otherwise. \textit{Recall@k:} Proportion of relevant items in the top-k results or the maximum number of correct answers. \textit{Mean Reciprocal Rank (MRR):} Reciprocal of the rank at which the first relevant item appears.

\begin{table}[t!]
    \centering
\begin{tabular}{c|c|c|c|c|c|c|c|c|}
\hline
STaRK & Metric & Dense	& QA-	& VSS & Multi- & VSS \&        & VSS \&    & 4Step    \\ 
Testset & & Retriever	& GNN	&     & VSS    & Claude3 Rer.  & GPT4 Rer. & Focus \\ \hline
\multirow{4}{*}{AMAZON} & Hit@1 &0.153&0.266&0.392&0.401&0.455&0.448&\textbf{0.476}\\ 
                           & Hit@5 &0.479&0.500&0.627&0.650&0.711&\textbf{0.712}	&0.676\\ 
&Recall@20	&0.445&	0.520&	0.533	&0.551	&0.538	&0.554		&\textbf{0.562} \\
                           & MRR	&0.302	&0.378	&0.504	&0.516	&0.559	&0.557	&	\textbf{0.565}\\ \hline
\multirow{4}{*}{MAG}    & Hit@1 &0.105	&0.129	  &0.290	&0.259	&0.365	&0.409		&\textbf{0.538}\\
                          & Hit@5	&0.352	&0.390	  &0.496	&0.504	&0.532	&0.582		&\textbf{0.692}\\ 
                          &Recall@20&	0.421&	0.470&0.483	&0.508	&0.484	&0.486		&\textbf{0.658}\\
                           &MRR&	0.213	&0.291	  &0.386	&0.369	&0.442	&0.490	&	\textbf{0.614}\\ \hline
\multirow{4}{*}{PRIME}  & Hit@1&	0.045	&  0.089	&0.126	&0.151	&0.178	&0.183&\textbf{0.393}\\ 
                            & Hit@5	&0.219	&  0.214	&0.315	&0.336	&0.369	&0.373&\textbf{0.532}\\
                            &Recall@20&	0.301&	0.296	&0.360	&0.381	&0.356	&0.341&\textbf{0.559}\\
                             &MRR	&0.124	&     0.147	&0.214	&0.235	&0.263	&0.266&\textbf{0.458}\\ \hline \hline             

\multirow{3}{*}{AMAZON} & Hit@1    & & & 0.395 &0.469 &0.560 &0.543 &\textbf{0.605}\\ 
                           & Hit@5 & -& -& 0.642 &\textbf{0.728} &\textbf{0.728} &\textbf{0.728 }&0.679\\ 
                           &  MRR  & & & 0.527 &0.587 &\textbf{0.644} &0.627 &0.643\\ \hline
\multirow{3}{*}{MAG}    & Hit@1    & & & 0.286 &0.238 &0.381 &0.369 &\textbf{0.476}\\
                          & Hit@5  & -& -& 0.417 &0.417 &0.464 &0.452 &\textbf{0.512}\\ 
                           & MRR   & & & 0.358 &0.314 &0.424 &0.403 &\textbf{0.492}\\ \hline
\multirow{3}{*}{PRIME}  & Hit@1    & & & 0.212 &0.257 &0.284 &0.284 &\textbf{0.505}\\ 
                        & Hit@5    &- &- & 0.404 &0.404 &0.477 &0.486 &\textbf{0.655}\\
                             & MRR & & & 0.298 &0.338 &0.362 &0.362& \textbf{0.579}\\ \hline        
\end{tabular}
    \caption{Test results on STaRK benchmark datasets. The upper three datasets are synthetically generated, the three datasets below are built by humans directly. For methods that use proprietary LLMs, only 10\% of synthetic test sets are used.}
\label{tab:stark_synth}
\end{table}

Table~\ref{tab:stark_synth} shows the results of our experiments in the benchmark environment. 4StepFocus strongly improves the results of all previous methods on all measured metrics on the medical, more relational PRIME dataset. The same applies to the MAG dataset, which quite evenly balances relational elements and those that require interpretation of unstructured data in its queries. 
On the AMAZON dataset, the hit rate of the first given answer of 4StepFocus still outperforms the other approaches, but pure VSS + Reranker and Multi-VSS achieve higher HIT@5 scores on both the human-built and the syntatically generated test sets. One explanation is that some triplets can be misleading. E.g., a triplet filtering ``green'' products might ignore all ``light green'' products, since only one out of 1,700 color nodes is chosen. This suggests the integration of VSS during the triplet substitution step as well in the future. 

\section{Conclusion and Future Work}
\label{sec:conclusion}
In this paper, we presented 4StepFocus, a pipeline that substantially improves answers of both LLMs but also other enhancing state-of-the art methods. Additionally, it includes external knowledge in a more traceable and reliable way. Our framework consists of multiple steps: identifying central entities and their relations in a user's query with an LLM, querying a domain-specific SKB for those information, using VSS among possible answers to take unstructured data into account in further prefiltering, and letting an LLM rerank the final results with background information given. Through experiments on a medical, a product recommendation, and an academic paper evaluation QA set, we demonstrated that our approach in contrast to graph embedding-based methods not only adds traceability, but also improves the overall precision of LLMs and makes them more reliable. While the addition to LLMs does not require pretraining, the need of embedded candidates is still a limitation to generalizability, which our approach does not solve yet. However, narrowing potential answers, with prefiltering consequently, can make holding embeddings of all general answer candidates obsolete.

There are several areas that can be explored to enhance our framework. For instance, the search of entities in a SKB could include vector similarity for a higher precision in the future. Further options are considering adjectives and predicates to capture more nuanced information, replacing pronouns with referenced objects/subjects for clarity, applying external reasoning taking quantifiers and negations into account, and combining multiple knowledge bases. By pursuing these avenues of future work, we can continue to enhance the capabilities of LLMs in accessing external knowledge and conducting autonomous reasoning for more evidence-based question answering. 


\begin{credits}
\subsubsection{\ackname} 
This work was part of the cluster for atherothrombosis and individualized medicine (curATime), funded by the German Federal Ministry of Education and Research (03ZU1202NA).

\end{credits}
%
%
%
\bibliographystyle{splncs04}
\bibliography{references}


\end{document}